\documentclass[10pt,twocolumn,letterpaper]{article}

\usepackage[pagenumbers]{cvpr} 

\usepackage{cuted} 

\usepackage{currfile} 

\usepackage{caption} 

\usepackage{booktabs}
\usepackage{multirow}
\usepackage{amsmath,amssymb}
\usepackage{xspace}
\newcommand{\method}{ClueRL\xspace}

\makeatletter\@namedef{ver@everyshi.sty}{}\makeatother
\usepackage{tikz}
\usetikzlibrary{positioning,arrows.meta,calc}
\usepackage{pifont}
\usepackage{tcolorbox}
\definecolor{findingbg}{RGB}{232,243,253}
\definecolor{findingframe}{RGB}{90,150,210}
\newtcolorbox{finding}{colback=findingbg, colframe=findingframe, boxrule=0.6pt, arc=2pt, left=6pt, right=6pt, top=4pt, bottom=4pt}
\definecolor{hintc}{HTML}{EB6834}
\usepackage{colortbl}
\colorlet{oursbg}{black!8}
\colorlet{upc}{black}\colorlet{dnc}{black}
\newcommand{\up}[1]{\,{\scriptsize\textcolor{upc}{+#1}}}
\newcommand{\dn}[1]{\,{\scriptsize\textcolor{dnc}{$-$#1}}}
\newcommand{\nod}{\,{\scriptsize\phantom{+0.0}}}

\definecolor{descc}{HTML}{2A78D6}
\colorlet{nonec}{black!45}

\definecolor{cvprblue}{rgb}{0.21,0.49,0.74}
\usepackage[pagebackref,breaklinks,colorlinks,allcolors=cvprblue]{hyperref}

\def\confName{CVPR}
\def\confYear{2027}

\title{Find the Clue, Not the Scene:\\Rethinking Reasoning for Entity Recognition in Vision-Language Models}

\author{
Kyungmin Min$^{1}$\footnotemark[1],
Soyeon Kim$^{2}$,
Junseok Kim$^{1}$,
Kyomin Jung$^{1,3}$\footnotemark[2] \\
$^{1}$IPAI, Seoul National University \quad
$^{2}$LG AI Research \quad
$^{3}$Dept. of ECE, Seoul National University \\
{\tt\small \{kyungmin97, kim.junseok, kjung\}@snu.ac.kr \quad kimcando@lgresearch.ai}
}

\begin{document}
\maketitle

\renewcommand{\thefootnote}{\fnsymbol{footnote}}
\footnotetext[1]{Work partially done during an internship at LG AI Research.}
\footnotetext[2]{Corresponding author.} 
\renewcommand{\thefootnote}{\arabic{footnote}}

\begin{abstract}

Answering a knowledge-based visual question requires first recognizing the entity in the image and then recalling knowledge about it.
Thinking vision-language models (VLMs) reason for thousands of tokens before answering, yet it remains underexplored how well such thinking VLMs perform each of these two steps.
We therefore analyze thinking VLMs on this task and make two key observations. (1) Recognition is the bottleneck, since 76.3\% of errors across four models on ViQuAE stem from misrecognizing the entity, and 84.9\% of these failures are resolved once the entity name is provided in the question. (2) Recognition is driven by entity-recall clues rather than by general descriptions of the image. An entity-recall clue is a distinctive visual detail that helps the model recall the entity, such as keys, a traditional attribute of Saint Peter.
Based on these findings, we propose \textbf{\method}, a reinforcement learning approach that extends Group Relative Policy Optimization (GRPO) with an advantage localized to the entity-recall clue in the thinking trace.
Rather than injecting new entity knowledge, ClueRL changes how VLMs think: it leads them to attend more closely to visual details and spot these clues, helping them use the knowledge they already have.
Consequently, it outperforms both GRPO and a perception-reward baseline on three entity recognition benchmarks. Moreover, this clue-seeking behavior generalizes beyond entity recognition, the only task used for training. Since recognition is the bottleneck, the gains carry over to knowledge-based visual question answering. They even extend to general multimodal reasoning, raising Qwen3.5-4B by 4.4 and 4.0 points on the ScienceQA and MMMU subsets of MMStar, respectively.

\end{abstract}

\newpage
\section{Introduction}
\label{sec:intro}

\input{fig/motivation}

Large language models (LLMs) store abundant factual knowledge in their parameters~\cite{petroni2019lama,kim-etal-2026-training}, yet they often fail to recall it, leading to factual errors~\cite{calderon2026emptyshelves}.
Thinking before answering helps LLMs recall this parametric knowledge~\cite{gekhman2026thinking,ma2026parametric}.
Such thinking is now standard in vision-language models (VLMs) as well~\cite{kimiteam2025kimivl,hong2025glm41v,qwen35,gemma4}.
However, factual questions posed to LLMs typically name the entity explicitly (\eg, ``When was the Eiffel Tower built?''), whereas visual questions often refer to it only through the image (\eg, ``When was this tower built?'').
A VLM must therefore first recognize the entity in the image and then recall knowledge about it.
It remains largely unexplored how well thinking VLMs perform each of these two steps.

\begin{figure}[t]
\centering
\includegraphics[width=\linewidth]{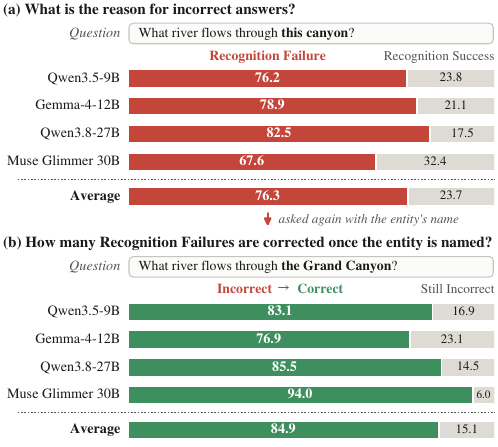}
\caption{\textbf{Entity recognition is the bottleneck} (ViQuAE, \%). (a) Incorrect answers are classified by whether the model's thinking identifies the correct entity \textbf{(Recognition Success)} or a different one \textbf{(Recognition Failure)}. (b) Percentage of Recognition Failures in (a) that are answered correctly when the image is replaced with the entity's name.}
\label{fig:bottleneck}
\end{figure}

To answer this question, we analyze four thinking VLMs (Qwen3.5~\cite{qwen35}, Qwen3.8~\cite{qwen38}, Gemma 4~\cite{gemma4}, and Muse Glimmer~\cite{museglimmer}) on ViQuAE~\cite{lerner2022viquae}, a knowledge-based visual question answering (VQA) benchmark whose questions require knowledge about the entity shown in the image.
We observe that, on average across the four models, 76.3\% of errors stem from misrecognizing the entity in the image.
Yet 84.9\% of these failures are resolved once the entity name is provided in the question (Figure~\ref{fig:bottleneck}).
These results indicate that entity recognition, rather than knowledge recall, is currently the primary bottleneck for thinking VLMs in knowledge-based VQA.

To understand what makes thinking effective for entity recognition, we examine how thinking traces differ between failed and successful recognition (Figure~\ref{fig:motivation}).
Both traces attend to the same object in the image, but the failed trace merely describes its general appearance, whereas the successful trace identifies an \textit{entity-recall clue}, a distinctive visual detail that helps the model recall the entity (\eg, spotting the \textit{Carrier Dome} leads to recognizing \textit{Syracuse University}).
To verify this observation quantitatively, we inject either an entity-recall clue or a general description of the image into the models' thinking traces and compare each with injecting nothing (Figure~\ref{fig:insert}).
Although both interventions add information to the thinking trace, the entity-recall clue improves entity recognition accuracy by 13.3 points over the no-injection baseline across the four VLMs, whereas the general description improves it by only 2.1 points.
Thus, what makes thinking effective for entity recognition is not describing more of the image, but finding a clue that brings the entity to mind.

Building on these findings, we propose \textbf{\method}, a reinforcement learning method that rewards the model for spotting entity-recall clues while thinking.
\method keeps the answer-correctness reward of Group Relative Policy Optimization (GRPO)~\cite{shao2024deepseekmath} and additionally assigns a localized advantage to the thinking tokens that state an entity-recall clue.
This advantage encourages the model to seek out such clues more often in its thinking.
Importantly, rather than injecting new entity knowledge, \method reshapes how the model thinks, helping it use the knowledge it already has.

We first evaluate \method on entity recognition, the task it is trained for.
On three entity recognition benchmarks, \method improves Qwen3.5-4B and Qwen3.5-9B over their base models by 1.9 and 2.1 points on average, respectively.
These are the largest gains among methods trained on the same data, including GRPO~\cite{shao2024deepseekmath} and Perception-R1~\cite{xiao2026perceptionr}, which rewards visual perception in general.
To confirm that these gains come from the intended change in thinking, we measure how often the thinking mentions an entity-recall clue.
On WorldVQA~\cite{worldvqa}, \method yields the highest clue mention rate among all methods (\eg, 63.2\% vs.\ 59.0\% for base Qwen3.5-4B; Table~\ref{tab:clue}), and its accuracy gains over the base model are larger on images that contain such clues.
This indicates that the models learn to spot entity-recall clues and recognize the entity from them.

Next, we ask whether this change in thinking also benefits tasks beyond entity recognition, the only task used for training.
Since recognition is the primary bottleneck of knowledge-based VQA, improving it should let the model apply the knowledge it already has.
Indeed, on three knowledge-based VQA benchmarks, Qwen3.5-4B and Qwen3.5-9B trained with \method improve over their base models by 2.0 and 1.6 points on average, respectively, without any training on knowledge questions.
Finally, the benefit extends even to general multimodal reasoning.
On MMStar~\cite{chen2024mmstar}, Qwen3.5-4B trained with \method gains 4.4 and 4.0 points over its base model on the questions drawn from ScienceQA~\cite{lu2022scienceqa} and MMMU~\cite{yue2024mmmu}, respectively, where many answers depend on fine visual details in scientific diagrams.
This suggests that closely inspecting visual details has become part of the model's thinking.

Our contributions are three-fold.
\begin{itemize}
    \item We show that entity recognition, rather than knowledge recall, is the primary bottleneck of thinking VLMs in knowledge-based VQA, and that recognition improves when their thinking includes an entity-recall clue rather than a general description of the image.
    \item Building on this analysis, we propose \textbf{\method}, which rewards the model for spotting entity-recall clues while thinking, via an advantage localized to the thinking tokens that state such a clue.
    \item \method achieves the highest entity recognition accuracy on three benchmarks among methods trained on the same data. The learned clue-seeking behavior also transfers to knowledge-based VQA and general multimodal reasoning without additional training.
\end{itemize}

\section{Related Work}
\label{sec:related}

\paragraph{Factual recall in VLMs.}
Benchmarks for factual recall in VLMs ask about an entity shown in an image and cover two tasks.
\textit{Entity recognition} is a single-hop task that asks the model to name the entity itself, such as a landmark or the brand of a logo~\cite{hu2023oven,worldvqa}.
\textit{Knowledge-based VQA} instead asks for knowledge about the entity, such as the CEO or founding year of that brand~\cite{lerner2022viquae,chen2023infoseek,mensink2023encyclopedic,cheng2025simplevqa,wang2025visualsimpleqa}.
This task thus takes two hops, entity recognition followed by knowledge recall.
Prior analyses study these two hops through the internal representations of non-thinking VLMs such as LLaVA and Qwen2.5-VL~\cite{cohen2025popvqa,ashok2025vlmrecall,venhoff2025toolate}.
Venhoff et al.~\cite{venhoff2025toolate} also evaluate chain-of-thought prompting, but only by its effect on final accuracy, which they find varies across models.
However, thinking VLMs internalize such reasoning through reinforcement learning~\cite{guo2025deepseekr1,kimiteam2025kimivl,hong2025glm41v}, and how well they perform the two tasks remains underexplored.
We therefore systematically analyze thinking VLMs on both tasks and study what makes their thinking effective.

\paragraph{Reinforcement learning in VLMs.}
Reinforcement learning is widely used to improve the reasoning of VLMs.
Outcome-based methods~\cite{huang2025visionr1,shen2025vlmr1} optimize models with GRPO~\cite{shao2024deepseekmath}, rewarding only final-answer correctness.
Perception-based methods additionally reward the visual information in a response, such as image descriptions generated before reasoning~\cite{xia2025visionaryr1,li2026visionsr,chen2025pebrr1,just2025perlvl} or coverage of visual statements from a teacher model~\cite{xiao2026perceptionr}.
These methods mainly target mathematical reasoning, and perception-based ones reward the visual content of a response as a whole.
For entity recognition, however, our analysis shows that general descriptions of the image help only marginally, whereas an entity-recall clue helps substantially.
\method therefore rewards only this clue, crediting the thinking tokens that state it rather than the whole response.

\suppressfloats[t]
\input{fig/cue_pipeline}

\section{What Limits Thinking VLMs in Knowledge-Based VQA?}
\label{sec:motivation}
\label{sec:diagnosis}

Knowledge-based VQA involves two hops, namely \textit{entity recognition}, which identifies the entity depicted in the image, and \textit{knowledge recall}, which recalls the fact about this entity that the question asks for.
In this section, we analyze thinking VLMs along these two hops.
Section~\ref{sec:diag_bottleneck} identifies which hop is the bottleneck, using an error analysis and a control experiment that varies whether the model must recognize the entity from the image or is given its name.
Section~\ref{sec:diag_hint} then intervenes on the models' thinking to investigate what kind of visual information helps them recognize the entity.

\subsection{Entity Recognition vs.\ Knowledge Recall}
\label{sec:diag_bottleneck}

We analyze four thinking VLMs, Qwen3.5-9B~\cite{qwen35}, Gemma-4-12B~\cite{gemma4}, Qwen3.8-27B~\cite{qwen38}, and Muse Glimmer 30B~\cite{museglimmer}, on ViQuAE~\cite{lerner2022viquae} (1,257 test questions), a knowledge-based VQA benchmark in which each question refers to an entity only through the image and asks for knowledge about it.

To locate where the models fail, we use an LLM judge (GPT-6 Luna~\cite{openai2026gpt6luna}) that reads each incorrect response, along with its thinking, and determines whether the model misrecognized the entity in the image as a different one.
We refer to such errors as \textit{Recognition Failures}.
On average across the four models, Recognition Failures account for 76.3\% of incorrect responses (Figure~\ref{fig:bottleneck}a).
Yet 84.9\% of these failures are resolved once the question names the entity instead of referring to it through the image (Figure~\ref{fig:bottleneck}b).
This suggests that most of their errors stem not from missing knowledge but from failing to recognize the entity in the image.

\begin{finding}
\textbf{Finding 1: Entity recognition is the bottleneck.}
Most errors of thinking VLMs in knowledge-based VQA stem from misrecognizing the entity rather than from missing knowledge; once the entity is named, the models answer most of these questions correctly.
\end{finding}

\begin{figure}[!t]
\centering
\includegraphics[width=\linewidth]{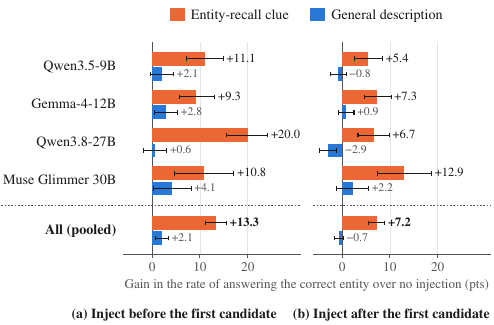}
\caption{\textbf{Entity-recall clues, not general descriptions, improve entity recognition.} An entity-recall clue or a general description is injected into an incorrect thinking trace, (a) right before or (b) right after its first candidate entity, and the model then resumes its thinking. Each bar shows how much the rate of answering the correct entity increases compared with injecting nothing. \textit{All} pools the questions of the four models, and error bars show 95\% bootstrap confidence intervals over questions.}
\label{fig:insert}
\end{figure}

\input{fig/method_v2}

\subsection{Entity-Recall Clues vs.\ General Descriptions}
\label{sec:diag_hint}

Given that entity recognition is the bottleneck, we next examine what kind of thinking leads the models to the correct entity.
Figure~\ref{fig:motivation} offers a hint.
Thinking that spots an \textit{entity-recall clue}, a distinctive visual feature that recalls the entity (\eg, keys for Saint Peter), reaches the correct entity.
Building on this observation, we test which kind of visual information in the thinking helps the model recognize the entity.

\paragraph{Experimental setting.}
We evaluate the first hop directly with an entity recognition task, \ie, entity identification questions (\eg, ``What is the name of the landmark in this image?'') on the images of ViQuAE~\cite{lerner2022viquae} (training/validation, 1,610 questions) and PopVQA~\cite{cohen2025popvqa} (2,446 questions), using the same four models as in Section~\ref{sec:diag_bottleneck}.
We construct entity-recall clues for each image in four steps (Figure~\ref{fig:cuepipe}).
GPT-6.1 Sol~\cite{openai2026gpt61sol} first proposes candidate clues given the image, the question, and the answer.
Three filters then follow.
A \textit{hallucination check} with GPT-6 Luna~\cite{openai2026gpt6luna} removes candidates that are not visible in the image, and a \textit{leakage check} removes candidates that reveal the answer, which would serve as a shortcut rather than a visual clue.
Finally, a \textit{decisiveness check} keeps a candidate only if GPT-6 Luna, given the clue without the image, recalls the correct entity.

\paragraph{Entity-recall clue injection.}
To identify which information in the thinking helps entity recognition, we inject different types of information into the models' thinking and compare the outcomes.
We use \textit{partially solved} questions, \ie, those answered correctly in 1 to 7 of 8 samples.
On these questions, the model possesses the knowledge required to recognize the entity, as it answers correctly in some samples.
Its failures can therefore be attributed to its thinking rather than to a lack of knowledge.
From each question, we take one thinking trace that ends in a wrong entity.
Using an LLM judge, we locate the first wrong candidate entity in this trace, \ie, the first entity that the thinking proposes as the answer.
We then inject an entity-recall clue, an additional general description of the image, or nothing as a baseline, either right before or right after this candidate.
Injecting before the candidate tests whether the information steers the thinking toward the correct entity, and injecting after it tests whether the information helps the model recover from the wrong candidate.
From each injection point, the model resumes its thinking 8 times, and an LLM judge decides whether the final answer is the correct entity.
As shown in Figure~\ref{fig:insert}, injecting a clue before the candidate raises entity recognition accuracy by 13.3 points over the no-injection baseline, whereas injecting a general description raises it by only 2.1 points.
Even when injected after the model has already named the incorrect candidate, the clue still helps it recover the correct entity, raising accuracy by 7.2 points, whereas the general description does not help ($-$0.7 points).
This gain from the clue holds at both positions in every model (Figure~\ref{fig:insert}).

\begin{finding}
\textbf{Finding 2: Entity-recall clues drive entity recognition.}
Unlike general descriptions of the image, which yield only a marginal gain, entity-recall clues substantially improve entity recognition and even help the model recover the correct entity after committing to an incorrect one.
\end{finding}

\section{\method}
\label{sec:method}

Building on insights from Section~\ref{sec:diagnosis}, we propose \method, which incentivizes the model to articulate entity-recall clues in its thinking.
The key idea is to credit the statement of a clue separately from the correctness of the answer.
As illustrated in Figure~\ref{fig:method}, \method proceeds in three steps: (i) the model thinks and then answers; (ii) an LLM judge scores the answer and locates the clue span in the thinking; and (iii) the clue reward is converted into an advantage that is added only to the clue span, on top of the GRPO advantage~\cite{shao2024deepseekmath}.
Below, we set up the problem, define the two rewards, and describe how they are combined in policy optimization.

\subsection{Problem Setup}
\label{sec:method_task}

Given an image $I$ and a question $q$ about the entity it shows, the policy $\pi_\theta$ generates a response $o$, whose $t$-th token we denote by $o_t$.
Each response consists of thinking followed by an answer, $o = [o^{\mathrm{think}}; o^{\mathrm{ans}}]$.
The thinking $o^{\mathrm{think}}$ is the text between the \texttt{<think>} and \texttt{</think>} tags, and the answer $o^{\mathrm{ans}}$ is the text that follows \texttt{</think>}.
We evaluate correctness only on $o^{\mathrm{ans}}$ and search for entity-recall clues only in $o^{\mathrm{think}}$.

\subsection{Outcome and Clue Rewards}
\label{sec:method_reward}

During training, the policy samples a group of $G$ responses (rollouts) $\{o_i\}_{i=1}^{G}$ for each question, and each rollout receives two rewards.

\paragraph{Outcome reward.}
An LLM judge compares $o_i^{\mathrm{ans}}$ with the target entity and sets $r_i = 1$ if the answer is correct and $r_i = 0$ otherwise.

\paragraph{Clue reward.}
Let $\mathcal{C}$ denote the set of entity-recall clues of a training question, constructed as described in Section~\ref{sec:diag_hint}.
Given $o_i^{\mathrm{think}}$ and $\mathcal{C}$, the LLM judge decides for each clue in $\mathcal{C}$ whether the thinking states it and, if so, extracts the span that first states it.
We set $c_i = 1$ if the thinking correctly states at least one clue and define its \textit{clue span} $\mathcal{S}_i$ as the tokens of the extracted spans; otherwise, $c_i = 0$ and $\mathcal{S}_i = \emptyset$.
For questions with $\mathcal{C} = \emptyset$, $c_i = 0$ for all rollouts.

\subsection{Localized Clue Advantage}
\label{sec:method_adv}

\method keeps the two rewards separate and converts each into its own advantage as in GRPO~\cite{shao2024deepseekmath}.
The \textit{outcome advantage} is the outcome reward normalized within the group,
\begin{equation}
\hat A_i = \frac{r_i - \mathrm{mean}(\{r_j\}_{j=1}^{G})}{\mathrm{std}(\{r_j\}_{j=1}^{G})},
\label{eq:grpo_adv}
\end{equation}
and the \textit{clue advantage} is the clue reward normalized in the same way,
\begin{equation}
\hat A^{c}_i = \frac{c_i - \mathrm{mean}(\{c_j\}_{j=1}^{G})}{\mathrm{std}(\{c_j\}_{j=1}^{G})}.
\label{eq:clue_adv}
\end{equation}
\method then adds the clue advantage to the outcome advantage only on the tokens of the clue span.
Thus, every token of a rollout receives the outcome advantage, while the tokens that state a clue additionally receive a localized clue advantage.
Formally, the advantage of the $t$-th token of rollout $i$ is
\begin{equation}
A_{i,t} =
\begin{cases}
\hat A_i + \lambda\, \dfrac{\bar L}{|\mathcal{S}_i|}\, \hat A^{c}_i & t \in \mathcal{S}_i, \\[6pt]
\hat A_i & \text{otherwise,}
\end{cases}
\label{eq:ecl}
\end{equation}
where $\lambda$ weights the clue advantage and $\bar L$ is the mean rollout length in the batch.
Since the clue span accounts for only a fraction of the thinking tokens, the remaining factor $\bar L / |\mathcal{S}_i|$ amplifies the clue advantage to give the span a strong training signal.

The policy is optimized with the GRPO objective using $A_{i,t}$,
\begin{equation}
\begin{aligned}
& \mathcal{J}(\theta) = \frac{1}{\sum_{i \in \mathcal{B}} |o_i|} \sum_{i \in \mathcal{B}} \sum_{t=1}^{|o_i|} \Big[ \min\!\big( \rho_{i,t} A_{i,t}, \\
& \quad \mathrm{clip}(\rho_{i,t}, 1{-}\epsilon, 1{+}\epsilon)\, A_{i,t} \big) - \beta\, D_{\mathrm{KL}}[\pi_\theta \,\|\, \pi_{\mathrm{ref}}] \Big],
\end{aligned}
\label{eq:objective}
\end{equation}
where $\rho_{i,t} = \pi_\theta(o_{i,t} \mid I, q, o_{i,<t}) / \pi_{\theta_{\mathrm{old}}}(o_{i,t} \mid I, q, o_{i,<t})$ is the importance ratio of the $t$-th token of rollout $i$ in the batch $\mathcal{B}$, $\epsilon$ is the clipping range, and $\beta$ weights the KL regularization toward the base model $\pi_{\mathrm{ref}}$.

\section{Experiments}
\label{sec:experiments}

\subsection{Experimental Setup}
\label{sec:exp_setup}

\paragraph{Training dataset.}
We train two thinking VLMs, Qwen3.5-4B and Qwen3.5-9B~\cite{qwen35}, on the entity recognition task of Section~\ref{sec:diag_hint}, \ie, entity identification questions (\eg, ``What is the name of the landmark in this image?'') on the images of ViQuAE~\cite{lerner2022viquae} (training/validation) and PopVQA~\cite{cohen2025popvqa}.
Images whose entity appears in any test set are removed.
During training, each model generates 8 rollouts per question, and GPT-6 Luna~\cite{openai2026gpt6luna} judges both rewards.
We apply LoRA~\cite{hu2022lora} (rank 16, $\alpha = 32$) to the language model, keep the vision encoder frozen, and train for 2 epochs.
For each model, we select the $\lambda$ that performs best on a held-out validation set, which gives $\lambda = 0.7$ for Qwen3.5-4B and $\lambda = 0.4$ for Qwen3.5-9B.

\begin{table}[t]
\centering
\small
\setlength{\tabcolsep}{4pt}
\resizebox{\columnwidth}{!}{%
\begin{tabular}{lcccc}
\toprule
Method & WorldVQA & SimpleVQA & ViQuAE & Overall \\
\midrule
\multicolumn{5}{l}{\textit{Qwen3.5-4B}} \\
Base & $17.5 \pm 0.5$ & $43.6 \pm 0.6$ & $36.6 \pm 0.3$ & $32.6$ \\
SFT & $17.7 \pm 0.7$ & $44.0 \pm 0.9$ & $38.4 \pm 0.7$ & $33.4$ \\
GRPO & $18.3 \pm 0.8$ & $43.9 \pm 0.3$ & $39.3 \pm 0.6$ & $33.8$ \\
Perception-R1 & $18.1 \pm 0.4$ & $45.0 \pm 0.7$ & $38.7 \pm 0.8$ & $33.9$ \\
\rowcolor{oursbg}
\method{} (ours) & $\mathbf{19.0 \pm 0.6}$ & $\mathbf{45.1 \pm 0.6}$ & $\mathbf{39.5 \pm 0.6}$ & $\mathbf{34.5}$ \\
\midrule
\multicolumn{5}{l}{\textit{Qwen3.5-9B}} \\
Base & $20.5 \pm 0.5$ & $44.6 \pm 0.3$ & $41.6 \pm 0.6$ & $35.6$ \\
SFT & $21.0 \pm 0.4$ & $44.8 \pm 0.7$ & $43.2 \pm 0.5$ & $36.3$ \\
GRPO & $21.1 \pm 0.6$ & $45.0 \pm 0.7$ & $44.4 \pm 0.8$ & $36.8$ \\
Perception-R1 & $21.0 \pm 0.8$ & $44.9 \pm 0.4$ & $43.7 \pm 0.6$ & $36.5$ \\
\rowcolor{oursbg}
\method{} (ours) & $\mathbf{22.3 \pm 0.8}$ & $\mathbf{45.4 \pm 0.7}$ & $\mathbf{45.4 \pm 0.7}$ & $\mathbf{37.7}$ \\
\bottomrule
\end{tabular}%
}
\caption{\textbf{Test accuracy (Avg@8) on the entity recognition task (1-hop).} Rewarding entity-recall clues \textbf{(\method)} outperforms rewarding only answer correctness (GRPO) or general perception (Perception-R1) on every benchmark for both models.}
\label{tab:main}
\end{table}

\paragraph{Benchmarks and evaluation.}
We evaluate on two tasks that cover the first hop and both hops of knowledge-based VQA, respectively.
The \textit{entity recognition task (1-hop)} asks for the entity shown in the image, and we use three benchmarks for it.
WorldVQA~\cite{worldvqa} evaluates atomic visual world knowledge by asking the model to name the entity in an image, with entities ranging from common to long-tail ones.
From SimpleVQA~\cite{cheng2025simplevqa}, we use the atomic questions, which ask for the entity in each image.
For ViQuAE, we construct an identification question for the image of each test question, which yields a 1-hop counterpart of each knowledge-based question.
The \textit{knowledge-based VQA task (2-hop)} asks for knowledge about the entity shown in the image.
For this task, we use the test questions of ViQuAE, SimpleVQA, and VisualSimpleQA~\cite{wang2025visualsimpleqa}.
For WorldVQA and SimpleVQA, we use the English questions.
Since training uses only identification questions, the 2-hop task measures whether improved entity recognition carries over to questions that require knowledge about the entity.
We report Avg@8, \ie, the accuracy averaged over 8 responses per question, together with the standard deviation over the 8 runs.
Correctness is judged by GPT-6 Luna~\cite{openai2026gpt6luna}.
Further details of training and evaluation, including the sampling settings and judge prompts, are provided in the supplementary material.

\paragraph{Baselines.}
We compare \method with the base model and three training methods that start from the same base model.
1)~\textbf{SFT} fine-tunes the model on its own correct responses to the training questions, including their thinking traces.
2)~\textbf{GRPO}~\cite{shao2024deepseekmath} uses only the outcome reward of Section~\ref{sec:method_reward} and is thus equivalent to \method without the clue advantage.
3)~\textbf{Perception-R1}~\cite{xiao2026perceptionr} additionally rewards visual perception.
Its perception reward is the fraction of reference visual statements that the response covers, where the references are extracted from the chain-of-thought responses of a teacher model.
This term is added to the outcome reward with a weight of 0.7, and an $n$-gram repetition penalty is applied as well.
We follow its original reward design and hyperparameters, but train it on our entity recognition data and use GPT-6.1 Sol~\cite{openai2026gpt61sol} as the teacher model.
Perception-R1 thus rewards visual perception in general, whereas \method rewards only entity-recall clues, the specific visual details that point to the entity.

\begin{table}[t]
\centering
\small
\setlength{\tabcolsep}{3.5pt}
\resizebox{\columnwidth}{!}{%
\begin{tabular}{lccc}
\toprule
& & \multicolumn{2}{c}{Accuracy} \\
\cmidrule(lr){3-4}
Method & Clue mention rate & Clue in image & No clue in image \\
\midrule
\multicolumn{4}{l}{\textit{Qwen3.5-4B}} \\
Base & $59.0$\nod & $25.9$\nod & $13.9$\nod \\
GRPO & $56.7$\dn{2.3} & $27.1$\up{1.2} & $14.4$\up{0.5} \\
Perception-R1 & $60.3$\up{1.3} & $26.5$\up{0.6} & $14.4$\up{0.5} \\
\rowcolor{oursbg}
\method{} (ours) & $\mathbf{63.2}$\up{4.2} & $\mathbf{28.2}$\up{2.3} & $\mathbf{14.9}$\up{1.0} \\
\midrule
\multicolumn{4}{l}{\textit{Qwen3.5-9B}} \\
Base & $62.2$\nod & $30.3$\nod & $16.3$\nod \\
GRPO & $63.1$\up{0.9} & $30.4$\up{0.1} & $17.0$\up{0.7} \\
Perception-R1 & $64.1$\up{1.9} & $30.5$\up{0.2} & $16.9$\up{0.6} \\
\rowcolor{oursbg}
\method{} (ours) & $\mathbf{66.7}$\up{4.5} & $\mathbf{32.6}$\up{2.3} & $\mathbf{17.8}$\up{1.5} \\
\bottomrule
\end{tabular}%
}
\caption{\textbf{Entity-recall clues at test time} (WorldVQA, \%). The clue mention rate is how often the thinking mentions an entity-recall clue on the questions whose image has one, and accuracy (Avg@8) is reported separately for the questions whose image has a clue and for those whose image has none. Small numbers are changes from Base.}
\label{tab:clue}
\end{table}
\subsection{Quantitative Results on Entity Recognition Tasks}
\label{sec:exp_main}

As shown in Table~\ref{tab:main}, \method outperforms all other methods on the entity recognition task for both Qwen3.5-4B and Qwen3.5-9B, improving overall accuracy over the base models by 1.9 and 2.1 points, respectively.
This improvement cannot be attributed to knowledge acquired about the test entities, as these entities are excluded from training.
Moreover, \method differs from GRPO only in the additional clue advantage.
Its margin over GRPO, which holds on every benchmark for both models, therefore shows that incentivizing entity-recall clues is effective for entity recognition.
In contrast, Perception-R1, which rewards visual perception in general, performs on par with GRPO and below \method.
This indicates that identifying entity-recall clues is more effective for entity recognition than improving general perception.

\begin{table}[t]
\centering
\small
\setlength{\tabcolsep}{4pt}
\resizebox{\columnwidth}{!}{%
\begin{tabular}{lcccc}
\toprule
Method & ViQuAE & SimpleVQA & VisualSimpleQA & Overall \\
\midrule
\multicolumn{5}{l}{\textit{Qwen3.5-4B}} \\
Base & $44.6 \pm 0.8$ & $41.3 \pm 0.3$ & $45.2 \pm 0.7$ & $43.7$ \\
SFT & $45.1 \pm 0.8$ & $41.5 \pm 0.1$ & $45.8 \pm 0.8$ & $44.1$ \\
GRPO & $46.9 \pm 0.9$ & $41.8 \pm 0.3$ & $46.8 \pm 1.0$ & $45.1$ \\
Perception-R1 & $47.6 \pm 0.6$ & $42.2 \pm 0.6$ & $46.6 \pm 1.2$ & $45.5$ \\
\rowcolor{oursbg}
\method{} (ours) & $\mathbf{47.9 \pm 0.5}$ & $\mathbf{42.4 \pm 0.7}$ & $\mathbf{46.9 \pm 0.9}$ & $\mathbf{45.7}$ \\
\midrule
\multicolumn{5}{l}{\textit{Qwen3.5-9B}} \\
Base & $55.9 \pm 1.1$ & $43.8 \pm 0.5$ & $51.2 \pm 0.8$ & $50.3$ \\
SFT & $56.3 \pm 1.0$ & $43.9 \pm 0.9$ & $52.0 \pm 0.8$ & $50.8$ \\
GRPO & $57.6 \pm 0.6$ & $43.7 \pm 0.7$ & $51.9 \pm 1.2$ & $51.0$ \\
Perception-R1 & $57.4 \pm 0.5$ & $43.9 \pm 0.8$ & $52.2 \pm 0.7$ & $51.2$ \\
\rowcolor{oursbg}
\method{} (ours) & $\mathbf{58.3 \pm 0.9}$ & $\mathbf{44.8 \pm 0.8}$ & $\mathbf{52.6 \pm 1.1}$ & $\mathbf{51.9}$ \\
\bottomrule
\end{tabular}%
}
\caption{\textbf{Test accuracy (Avg@8) on the knowledge-based VQA task (2-hop).} \method achieves the highest accuracy on all three benchmarks for both models.}
\label{tab:transfer}
\end{table}

\subsection{Does \method Change How the Model Thinks?}
\label{sec:exp_clue}
We next examine whether this improvement stems from a change in the model's thinking behavior.
Specifically, we ask whether the model mentions entity-recall clues more often and whether its accuracy gain arises from such clue-based thinking.
To this end, we annotate entity-recall clues on the WorldVQA test images using the pipeline of Section~\ref{sec:diag_hint}, which yields at least one clue for 30.4\% of the questions.
An LLM judge determines whether the thinking mentions a clue.
We then compare the accuracy gains on questions with and without a clue in the image.

As shown in Table~\ref{tab:clue}, \method yields the largest increase in the clue mention rate for both models, whereas GRPO even decreases it for Qwen3.5-4B.
This indicates that rewarding the answer alone does not encourage the model to mention clues.
Furthermore, the gain of \method over the base model is larger on questions whose images contain a clue (+2.3 points for both models) than on those without one (+1.0 and +1.5 points).
Taken together, these results indicate that \method improves entity recognition through clue-based thinking.
In contrast, Perception-R1 shows that a higher clue mention rate alone does not guarantee better recognition.
For both models, Perception-R1 mentions clues more often than GRPO (+3.6 and +1.0 points), yet its accuracy on questions whose images contain a clue is lower or comparable ($-$0.6 and +0.1 points).
Unlike \method, Perception-R1 rewards all reference visual statements equally, including general descriptions that contribute little to recognition (Section~\ref{sec:diag_hint}).
We hypothesize that distributing the reward across many such statements dilutes the credit assigned to clue-based thinking, so that the model mentions the clue without relying on it to determine the entity.

\begin{figure}[t]
\centering
\includegraphics[width=\linewidth]{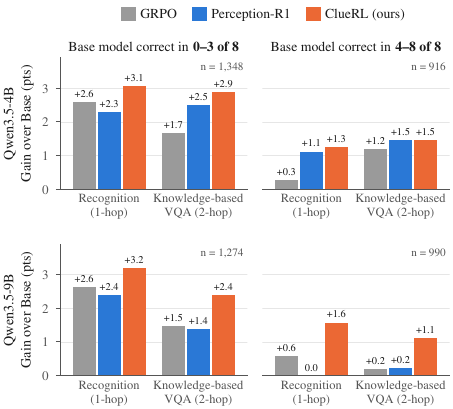}
\caption{\textbf{Knowledge-based VQA gains follow recognition gains.} Gains over the base model in the low-recognition (left) and high-recognition (right) groups, defined by the number of correct recognition answers of the base model out of 8. $n$ denotes the number of images in each group.}
\label{fig:hoplink}
\end{figure}

\subsection{Transfer to Knowledge-Based and General VQA}
\label{sec:exp_transfer}

Thus far, we have shown that the bottleneck of knowledge-based VQA lies in its first hop, entity recognition (Section~\ref{sec:diagnosis}), and that \method improves this hop by training the model to find entity-recall clues and reason from them (Sections~\ref{sec:exp_main} and~\ref{sec:exp_clue}).
We now examine two forms of transfer.
First, since \method is trained only on the first hop, we test whether improving it alone also improves knowledge-based VQA.
Second, we test whether this clue-based thinking also benefits general multimodal tasks.

\input{fig/mmstar_case}

\paragraph{Knowledge-based VQA.}
As shown in Table~\ref{tab:transfer}, the models trained with \method achieve the highest accuracy on every knowledge-based VQA benchmark, improving overall accuracy over the base models by 2.0 and 1.6 points.
We next examine whether this gain indeed comes from better entity recognition.
For this analysis, we use the 2,264 images of ViQuAE and SimpleVQA, each paired with both an entity recognition question and a knowledge-based question about the same entity, so that gains in recognition and in knowledge-based VQA can be compared on the same images.
We divide these images into a low-recognition group, whose entities the base model recognizes in 0--3 of its 8 answers, and a high-recognition group, whose entities it recognizes in 4--8.
Figure~\ref{fig:hoplink} shows how much \method and the two baselines improve over the base model in each group.
In both groups, \method improves recognition the most, and its knowledge-based accuracy rises along with it.
Interestingly, in the high-recognition group of Qwen3.5-9B, GRPO and Perception-R1 hardly improve knowledge-based accuracy (+0.2 points), whereas \method improves it by 1.1 points.
This suggests that \method recognizes these entities more consistently across samples.
Indeed, the number of images whose entity Qwen3.5-9B recognizes in all 8 samples rises from 693 to 794 (out of 2,264) with \method, exceeding both GRPO (750) and Perception-R1 (756).
We hypothesize that stating entity-recall clues makes recognition more stable, which in turn improves knowledge-based VQA.
Overall, these results indicate that improving entity recognition carries over to knowledge-based VQA, as it allows the model to apply the knowledge recall ability it already possesses.

\begin{table}[t]
\centering
\small
\setlength{\tabcolsep}{3.5pt}
\resizebox{\columnwidth}{!}{%
\begin{tabular}{lccccccc}
\toprule
Method & AI2D & MMBench & MMMU & MathVista & SEEDBench & ScienceQA & Avg. \\
\midrule
\multicolumn{8}{l}{\textit{Qwen3.5-4B}} \\
Base & $\mathbf{64.8}$ & $85.7$ & $68.4$ & $86.2$ & $56.9$ & $79.3$ & $73.5$ \\
GRPO & $63.7$ & $\mathbf{85.9}$ & $68.2$ & $86.4$ & $\mathbf{57.2}$ & $77.4$ & $73.1$ \\
\rowcolor{oursbg}
\method{} & $64.3$ & $85.7$ & $\mathbf{72.4}$ & $\mathbf{87.3}$ & $56.7$ & $\mathbf{83.7}$ & $\mathbf{75.0}$ \\
\rowcolor{oursbg}
\quad{\footnotesize $\Delta$ vs.\ Base} & {\footnotesize\textcolor{dnc}{$-$0.5}} & {\footnotesize\textcolor{black}{$\pm$0.0}} & {\footnotesize\textcolor{upc}{+4.0}} & {\footnotesize\textcolor{upc}{+1.1}} & {\footnotesize\textcolor{dnc}{$-$0.2}} & {\footnotesize\textcolor{upc}{+4.4}} & {\footnotesize\textcolor{upc}{+1.5}} \\
\midrule
\multicolumn{8}{l}{\textit{Qwen3.5-9B}} \\
Base & $69.1$ & $88.1$ & $72.4$ & $\mathbf{88.5}$ & $\mathbf{59.7}$ & $84.6$ & $77.1$ \\
GRPO & $70.0$ & $88.2$ & $71.5$ & $88.2$ & $58.1$ & $86.4$ & $77.1$ \\
\rowcolor{oursbg}
\method{} & $\mathbf{71.4}$ & $\mathbf{88.3}$ & $\mathbf{73.1}$ & $88.2$ & $58.3$ & $\mathbf{88.8}$ & $\mathbf{78.0}$ \\
\rowcolor{oursbg}
\quad{\footnotesize $\Delta$ vs.\ Base} & {\footnotesize\textcolor{upc}{+2.3}} & {\footnotesize\textcolor{upc}{+0.2}} & {\footnotesize\textcolor{upc}{+0.7}} & {\footnotesize\textcolor{dnc}{$-$0.3}} & {\footnotesize\textcolor{dnc}{$-$1.4}} & {\footnotesize\textcolor{upc}{+4.2}} & {\footnotesize\textcolor{upc}{+0.9}} \\
\bottomrule
\end{tabular}%
}
\caption{\textbf{General capability on MMStar} (Avg@8, \%), by the source benchmark of its questions. $\Delta$ is the change of \method from Base.}
\label{tab:mmstar}
\end{table}

\paragraph{General multimodal benchmark.}
We further evaluate on MMStar~\cite{chen2024mmstar}, a general multimodal benchmark whose questions are curated from multiple existing benchmarks and cannot be answered without the image.
Since these sources cover diverse capabilities, from perception to reasoning, MMStar allows us to assess how broadly the effect of \method extends.
As shown in Table~\ref{tab:mmstar}, the models trained with \method achieve the highest overall accuracy for both base models, indicating that training on entity recognition does not compromise general capability.
The gains are most pronounced on the questions drawn from ScienceQA and MMMU, where Qwen3.5-4B trained with \method improves over the base model by 4.4 and 4.0 points, respectively.
Many of these questions involve scientific diagrams, in which the answer often depends on a fine visual detail, such as the relative size of two objects, a label attached to one part, or the way the parts are connected.
Figure~\ref{fig:mmstar_case} shows two such cases.
In (a), the base model interprets the magnets in the two pairs as identical and concludes that the forces are equal, whereas our model notices that the magnet in Pair 2 is larger and answers correctly.
In (b), the GRPO baseline reads only the first entry in the adjacency list of V1 and treats it as the only neighbor, whereas our model reads the entire list, finds V3 and V2 as well, and obtains the correct traversal.
These cases suggest that the tendency to seek a decisive visual detail, learned on entity recognition, carries over to such questions.

\section{Conclusion}
\label{sec:conclusion}

In this paper, we examine how well thinking VLMs perform each step of knowledge-based VQA.
We find that entity recognition, rather than knowledge recall, is the primary bottleneck, as most recognition failures are resolved once the entity is named.
Moreover, what drives recognition is not describing more of the image but finding an entity-recall clue in it.
Based on these findings, we propose \textbf{\method}, which extends GRPO to reward the model for spotting entity-recall clues, with an advantage localized to the thinking tokens where such a clue appears.
Trained with \method, Qwen3.5-4B and Qwen3.5-9B spot entity-recall clues more often in their thinking and achieve the highest entity recognition accuracy among the compared methods.
These gains transfer to knowledge-based VQA and a general multimodal benchmark, despite training only on entity recognition.

\section*{LLM Usage Statement}

We used ChatGPT (OpenAI) and Claude (Anthropic) to revise the manuscript and refine the writing style. We used Claude Code (Anthropic) to write experiment code and generate figures. All LLM-generated content was reviewed by the authors.

{
    \small
    \bibliographystyle{ieeenat_fullname}
    \bibliography{main}
}

\clearpage
\clearpage
\maketitlesupplementary

\appendix
\newtcolorbox{promptblock}[1][]{colback=black!5, colframe=black!65, coltext=black, coltitle=white, fonttitle=\footnotesize, fontupper=\scriptsize, boxrule=0.5pt, arc=2pt, boxsep=0pt, left=5pt, right=5pt, top=4pt, bottom=4pt, toptitle=2.5pt, bottomtitle=2.5pt, before skip=4pt, after skip=7pt, before upper={\setlength{\topsep}{0pt}\setlength{\partopsep}{0pt}}, #1}
\section{Experimental Setup}
\label{sec:supp_setup}

\paragraph{Models.}
We use Qwen3.5-4B, Qwen3.5-9B~\cite{qwen35}, Gemma-4-12B~\cite{gemma4}, Qwen3.8-27B~\cite{qwen38}, and Muse Glimmer 30B~\cite{museglimmer} in thinking mode, with their default chat templates and no system prompt.
We set the reasoning level of Qwen3.8-27B and Muse Glimmer 30B to medium and generate all responses with vLLM~\cite{kwon2023vllm}.

\paragraph{Thinking budget control.}
We limit the thinking to 4,096 tokens.
If the thinking does not end within its limit, we cut it there, append the end-of-thinking tag, and let the model answer.
We judge only the answer, \ie, the text after the thinking.

\paragraph{Temperature.}
We use a temperature of 0.6 and a top-$p$ of 0.95 for the Qwen models, and the recommended setting (temperature 1.0, top-$p$ 0.95, top-$k$ 64) for Gemma-4-12B and Muse Glimmer 30B.

\section{Ablation on the Localized Clue Advantage}
\label{sec:supp_ablation}

We ablate the localized clue advantage of \method on Qwen3.5-4B with $\lambda = 0.7$.

\paragraph{Notation.}
For each training question, the policy generates $G$ responses, which we call rollouts.
We denote the $i$-th rollout by $o_i$ and its number of tokens by $|o_i|$.
The outcome reward $r_i$ is 1 if the answer of $o_i$ is correct and 0 otherwise.
The clue reward $c_i$ is 1 if the thinking of $o_i$ states an entity-recall clue of the image and 0 otherwise.
The clue span $\mathcal{S}_i$ is the set of thinking tokens that state the clue, and it is empty when $c_i = 0$.
We write $A_{i,t}$ for the advantage of the $t$-th token of $o_i$ and $\bar L$ for the mean number of tokens per rollout in the batch.

\begin{table}[t]
\centering
\footnotesize
\setlength{\tabcolsep}{3.5pt}
\begin{tabular}{@{}lccccc@{}}
\toprule
 & Separate & Clue & Span & Clue & Val. \\
Variant & norm. & span & scaling & mention & acc. \\
\midrule
GRPO & -- & -- & -- & 56.6 & 44.4 \\
(a) Scalar clue reward & & & & 55.8 & 42.4 \\
(b) Without localization & \checkmark & & \checkmark & 57.7 & 43.1 \\
(c) Without span scaling & \checkmark & \checkmark & & 57.7 & 44.4 \\
\rowcolor{oursbg} \method & \checkmark & \checkmark & \checkmark & \textbf{67.1} & \textbf{47.9} \\
\bottomrule
\end{tabular}
\caption{\textbf{Ablation of the localized clue advantage} (Qwen3.5-4B, $\lambda = 0.7$, \%). Clue mention is how often the thinking of a training rollout states a clue, on the training questions with a clue in the second epoch. Val.\ acc.\ is the accuracy of the final model on a validation set held out from the training set (8\% of the questions), averaged over 4 samples per question.}
\label{tab:supp_ablation}
\end{table}

\paragraph{\method.}
\method normalizes the two rewards separately over the $G$ rollouts of a question,
\begin{equation}
\begin{aligned}
\hat A_i &= \frac{r_i - \mathrm{mean}(\{r_j\}_{j=1}^{G})}{\mathrm{std}(\{r_j\}_{j=1}^{G})}, \\
\hat A^{c}_i &= \frac{c_i - \mathrm{mean}(\{c_j\}_{j=1}^{G})}{\mathrm{std}(\{c_j\}_{j=1}^{G})},
\end{aligned}
\label{eq:supp_norm}
\end{equation}
which gives the outcome advantage $\hat A_i$ and the clue advantage $\hat A^{c}_i$.
It then adds the clue advantage, weighted by $\lambda$, only on the clue span,
\begin{equation}
A_{i,t} =
\begin{cases}
\hat A_i + \lambda\, \dfrac{\bar L}{|\mathcal{S}_i|}\, \hat A^{c}_i & t \in \mathcal{S}_i, \\[6pt]
\hat A_i & \text{otherwise,}
\end{cases}
\label{eq:supp_clue}
\end{equation}
which is Main Paper Eq.~3.
This advantage has three components, namely the separate normalization of the clue reward, the localization on the clue span, and the scaling by $\bar L / |\mathcal{S}_i|$.
We train three variants that remove these components, with all other settings unchanged.

\paragraph{(a) Scalar clue reward.}
This variant removes all three components.
It adds the clue reward to the outcome reward and normalizes the sum,
\begin{equation}
\tilde r_i = r_i + \lambda c_i, \quad
A_{i,t} = \frac{\tilde r_i - \mathrm{mean}(\{\tilde r_j\}_{j=1}^{G})}{\mathrm{std}(\{\tilde r_j\}_{j=1}^{G})}.
\label{eq:abl_scalar}
\end{equation}
Unlike \method, the clue reward is not normalized separately, and every token of a rollout receives the same advantage.

\paragraph{(b) Without localization.}
This variant keeps the separate normalization but spreads the clue advantage over all $|o_i|$ tokens of a rollout that states a clue,
\begin{equation}
A_{i,t} = \hat A_i + \lambda\, \frac{\bar L}{|o_i|}\, c_i\, \hat A^{c}_i \quad \text{for all } t.
\label{eq:abl_all}
\end{equation}
Summed over a rollout, the clue advantage is $\lambda \bar L \hat A^{c}_i$ as in \method, so only its placement differs.

\paragraph{(c) Without span scaling.}
This variant keeps the separate normalization and the localization but removes the factor $\bar L / |\mathcal{S}_i|$,
\begin{equation}
A_{i,t} =
\begin{cases}
\hat A_i + \lambda\, \hat A^{c}_i & t \in \mathcal{S}_i, \\[3pt]
\hat A_i & \text{otherwise.}
\end{cases}
\label{eq:abl_noscale}
\end{equation}
Each token of the clue span thus receives a clue advantage that is $\bar L / |\mathcal{S}_i|$ times smaller than in \method.

\paragraph{Results.}
We compare the three variants with GRPO and \method (Table~\ref{tab:supp_ablation}).
The three variants leave the clue mention rate close to that of GRPO (55.8--57.7\% against 56.6\%), and their validation accuracy is likewise comparable to GRPO (42.4--44.4\% against 44.4\%).
In contrast, \method raises the clue mention rate to 67.1\% and improves the validation accuracy to 47.9\%.
Both the localization and the span scaling are therefore necessary for the clue reward to take effect.

\section{Ablation on the Clue Advantage Weight $\lambda$}
\label{sec:supp_lambda}

\begin{figure}[t]
\centering
\includegraphics[width=\linewidth]{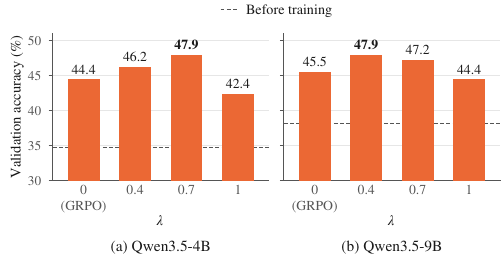}
\caption{\textbf{Ablation on $\lambda$.} Validation accuracy of the final model for each $\lambda$, where $\lambda = 0$ corresponds to GRPO. The dashed line is the accuracy before training.}
\label{fig:supp_lambda}
\end{figure}

We train \method with $\lambda \in \{0.4, 0.7, 1\}$ and compare the validation accuracy of the final models (Figure~\ref{fig:supp_lambda}).
The validation accuracy is highest at $\lambda = 0.7$ for Qwen3.5-4B and at $\lambda = 0.4$ for Qwen3.5-9B.
We use these values in the main experiments.

\section{Analysis of the Clue Mention Rate}
\label{sec:supp_training}

\begin{figure}[t]
\centering
\includegraphics[width=\linewidth]{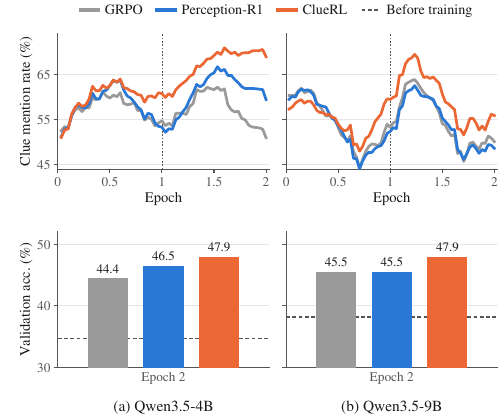}
\caption{\textbf{Analysis of the clue mention rate.} Clue mention rate of the training rollouts (top, moving average over 15 steps) and validation accuracy of the final model (bottom), both measured as in Table~\ref{tab:supp_ablation}. All methods see the same batches in the same order in both epochs, so the curves share the same pattern. The dashed line is the accuracy before training.}
\label{fig:supp_training}
\end{figure}

We observe how the clue mention rate of the training rollouts changes during training for GRPO, Perception-R1, and \method (Figure~\ref{fig:supp_training}).
The three methods start from similar clue mention rates.
As training proceeds, \method keeps a higher clue mention rate than the two baselines, averaging 67.1\% on Qwen3.5-4B and 61.4\% on Qwen3.5-9B in the second epoch, whereas GRPO averages 56.6\% and 55.4\%.
\method also achieves the highest validation accuracy on both models.
The higher clue mention rate carries over to the test images, whose entities never appear in the training data.
Specifically, on WorldVQA, \method raises the clue mention rate of the base model from 59.0\% to 63.2\% on Qwen3.5-4B and from 62.2\% to 66.7\% on Qwen3.5-9B, the largest increase among all methods (Main Paper Table~2).
This suggests that \method does not memorize the clues of the training images but learns a thinking behavior that finds entity-recall clues in images of unseen entities.

\input{fig/supp_case}

\section{Entity Recognition Questions for ViQuAE}
\label{sec:supp_hops}

ViQuAE~\cite{lerner2022viquae} is a knowledge-based VQA benchmark whose questions refer to an entity only through the image and ask for knowledge about it.
It labels each image with the Wikidata entity it shows but has no question about the entity itself.
We therefore add an identification question on the same image and use the labeled entity as its answer.
We choose the question by the type of the entity, \eg, ``What is the name of the place or building in this image?'' for a place or building, and GPT-6 Luna labels the type from the entity's name and the ViQuAE question.
This gives 1,257 test pairs, each with an entity recognition question (1-hop) and a knowledge-based VQA question (2-hop) on the same image.

\section{Case Study on Entity Recognition}
\label{sec:supp_case}

We compare the responses of Qwen3.5-9B to an entity recognition question of WorldVQA before training (Base) and after training with GRPO, Perception-R1, and \method (Figure~\ref{fig:supp_case}).
The question, its image, and the entity it shows are not seen during training.
The image has an entity-recall clue, the southern basin divided into rectangular evaporation ponds.
Base, GRPO, and Perception-R1 judge the lake by its elongated shape, take the ponds for a channel system or a solar farm, and answer the Sea of Galilee.
\method identifies the evaporation ponds, treats them as a clue, and recognizes the Dead Sea.

\section{Evaluation Settings}
\label{sec:supp_eval}

\paragraph{Benchmarks.}
For the entity recognition task, we use the 1,874 questions of WorldVQA~\cite{worldvqa}, the 1,007 atomic questions of SimpleVQA~\cite{cheng2025simplevqa}, and the 1,257 identification questions of ViQuAE.
For the knowledge-based VQA task, we use the 1,257 questions of the official test split of ViQuAE, the 1,007 original questions of SimpleVQA, and the 500 questions of VisualSimpleQA~\cite{wang2025visualsimpleqa}.
We use only the English questions of WorldVQA and SimpleVQA.
As a general multimodal benchmark, we use MMStar~\cite{chen2024mmstar}, which contains 1,500 multiple-choice questions drawn from six source benchmarks, namely AI2D, MMBench, MMMU, MathVista, SEEDBench, and ScienceQA.

\paragraph{Judges.}
We judge correctness with GPT-6 Luna, using the judge prompts in Figure~\ref{fig:supp_judges}.
We use the prompt in (a) for WorldVQA, SimpleVQA, and the identification questions of ViQuAE.
We use the prompt in (b) for the 2-hop questions of ViQuAE and for VisualSimpleQA.
For MMStar, we use VLMEvalKit~\cite{duan2024vlmevalkit}.
We measure the clue mention rate with the judge prompt in Figure~\ref{fig:supp_clue_judge}, using only the ``hints'' field of its output.

\section{Training Details}
\label{sec:supp_train}

We use the same hyperparameters for GRPO, Perception-R1, and \method (Table~\ref{tab:supp_hparams}).
We apply LoRA to the language model and keep the vision encoder frozen.
Each step samples 8 rollouts for each of 27 questions.
We run the experiments primarily on four NVIDIA A100 GPUs.

\begin{table}[t]
\centering
\footnotesize
\setlength{\tabcolsep}{4pt}
\begin{tabular}{@{}ll@{}}
\toprule
Hyperparameter & Value \\
\midrule
LoRA rank / $\alpha$ & 16 / 32 \\
LoRA modules & attention and MLP projections \\
Vision encoder & frozen \\
Learning rate & $5 \times 10^{-5}$, constant \\
Weight decay & 0 \\
Questions per step & 27 \\
Rollouts per question $G$ & 8 \\
Rollout sampling & temperature 0.6, top-$p$ 0.95 \\
Clipping range $\epsilon$ & 0.2 \\
KL weight $\beta$ & 0.01 \\
Epochs & 2 \\
\bottomrule
\end{tabular}
\caption{\textbf{Training hyperparameters}, shared by GRPO, Perception-R1, and \method.}
\label{tab:supp_hparams}
\end{table}

\section{Discussion}
\label{sec:supp_discussion}

\paragraph{Clues and parametric knowledge.}
The entity-recall clues are generated by GPT-6.1 Sol, and a clue passes the decisiveness check only if GPT-6 Luna recalls the entity from it (Figure~\ref{fig:supp_clue_prompts}).
We use these models to draw on their extensive world knowledge, since deciding which visual details point to an entity requires knowing the entity.
The clues are used only during training, to reward the thinking of the model, so whether the model can recall the entity from a clue depends on its own parametric knowledge.
One might thus expect little gain on compact models, yet \method improves entity recognition on Qwen3.5-4B and Qwen3.5-9B for entities not seen during training, which indicates that even these models hold knowledge that their default thinking leaves unused.
We therefore expect \method to be more effective on models with more parametric knowledge, for which a clue more readily evokes the entity.

\paragraph{A complementary view of entity recognition.}
WorldVQA~\cite{worldvqa} finds that the difficulty of entity recognition follows the rarity of the entity and attributes it to scarce knowledge of long-tail entities rather than to visual ambiguity.
It accordingly points to more diverse pre-training data as the way forward, which widens the knowledge of the model so that fewer entities fall outside it.
\method does not aim to widen this knowledge.
Instead, it improves entity recognition with the knowledge of the model largely unchanged, by changing the thinking behavior so that the model makes fuller use of what it already knows.
The gains on test images whose entities never appear in the training data (Main Paper Table~1) support this view, since they cannot come from newly memorized entities.
The two directions are therefore complementary: wider pre-training gives the model more knowledge, and \method helps the model use it during thinking.

\paragraph{Limitations.}
\method relies on an external LLM judge for the clue reward.
A rule-based check with exact matching would remove this dependence, but it would be too strict, since the thinking can state the same clue in different words.
We therefore use an LLM judge, which decides from the context whether the clue is stated.
In addition, \method targets entity-recall clues, so its effect can be smaller when the image contains no such clue.
To examine this effect, Main Paper Table~2 compares images with and without a clue.
While the gain of \method is larger on images with a clue, \method still improves accuracy more than the baselines on images without a clue.
This suggests that \method leads the model to inspect finer visual details in its thinking, even when no entity-recall clue is present.
Finally, ClueRL does not explicitly prevent the model from hallucinating such details,
which remains an open problem for VLMs~\cite{min2025mitigating, min2026reflectcap, li2023evaluating,favero2024multi,huang2024opera}.

\paragraph{Ethical considerations.}
Open-weight models in our study are obtained from their official releases, and proprietary models are accessed through their official APIs, in accordance with their licenses and terms of use.
All data come from public benchmarks released for research, and we collect no new images or personal data.
Since \method improves entity recognition, it could also make it easier to identify people in images.
Any use of \method for identifying people should therefore respect privacy and applicable regulations.

\section{Prompt Templates}
\label{sec:supp_prompts}

We present the prompt templates in the order of the judge prompts for answer correctness (Figure~\ref{fig:supp_judges}), the judge prompt for the clue mention rate (Figure~\ref{fig:supp_clue_judge}), the judge prompt for Recognition Failure and Recognition Success (Figure~\ref{fig:supp_recog_judge}), and the prompts for constructing entity-recall clues (Figure~\ref{fig:supp_clue_prompts}).

\onecolumn

\begin{center}
\begin{minipage}{0.98\textwidth}
\captionsetup{type=figure}
\begin{promptblock}[title={\textbf{(a) Judge prompt for WorldVQA, SimpleVQA, and the identification questions of ViQuAE}}]
\begin{verbatim}
### Role
You are an expert judge specialized in evaluating the correctness of answers. Your task is to assess whether
a model-generated answer is correct based on a given question, the model's response, and the ground truth
answer.

### Task: Evaluate Answer Correctness
Please classify the model's response into one of the following three categories. Ignore differences in
formatting, punctuation, language (Chinese vs. English), or abbreviations/full names. Focus strictly on the
**core semantics** and the **level of detail (granularity)**:

1. **Correct**:
    - The model answer contains the core information of the ground truth.
    - The model answer is semantically consistent with the ground truth and contains no contradictions.
    - **The granularity of the model answer is equal to or finer than the ground truth.**
    - Extra irrelevant information is allowed as long as it does not conflict with the ground truth.

2. **Incorrect**:
    - The model answer provides information that contradicts the ground truth.
    - The model answer provides the wrong specific entity, value, or description.
    - **The granularity of the model answer is coarser than the ground truth**, leading to incomplete or
insufficiently specific information.
    - Even if the model expresses uncertainty but follows up with a wrong answer (e.g., "I'm not sure, maybe
it's B" when the truth is A), it is considered Incorrect.

3. **Unattempted**:
    - The model explicitly states it does not know the answer (e.g., "I don't know," "I cannot answer this
question").
    - The model suggests the user search elsewhere (e.g., "Please search the internet").
    - The model answer contains no information from the ground truth but provides no incorrect or
contradictory information.

### Output Format
Please strictly follow this two-line format for your output:
1. **Evaluation**: [A brief explanation of your reasoning]
2. **Label**: [Final classification: "Correct", "Incorrect", or "Unattempted"]
\end{verbatim}
\par\vspace{3pt}\textit{(The six examples of this prompt are omitted here.)}\par\vspace{3pt}
\begin{verbatim}
### Current Task
Input:
'''
Question: {question}
Model Answer: {model_answer}
Ground Truth Answer: {ground_truth_answer}
'''

Evaluation:
\end{verbatim}
\end{promptblock}
\begin{promptblock}[title={\textbf{(b) Judge prompt for the 2-hop questions of ViQuAE and VisualSimpleQA}}]
\begin{verbatim}
You are grading an answer to a question about the entity shown in an image.

Question: {question}
Ground truth (any one option is enough): {reference}
Model answer: {candidate}

Correct: the model's final answer matches one of the ground truth options. Spelling, wording, or language
differences are fine, and so is a more specific answer that contains an option (e.g. "Munich, Germany" for
"Germany").
Incorrect: a different or vaguer answer, several conflicting answers, or no committed answer.
Use only the ground truth options, not your own knowledge.

Reply with one word: Correct or Incorrect.
\end{verbatim}
\end{promptblock}
\caption{\textbf{Judge prompts for answer correctness.}}
\label{fig:supp_judges}
\end{minipage}
\end{center}

\begin{center}
\begin{minipage}{0.98\textwidth}
\captionsetup{type=figure}
\begin{promptblock}[title={\textbf{Prompt}}]
\begin{verbatim}
Below is a QUESTION about an image, the TARGET entity that the image actually shows, a list of visual HINTS
that are present in the image, and the REASONING a model wrote about the image before answering.
Return JSON only, with these fields:
"first_candidate": the FIRST place in the reasoning where the model proposes a specific candidate for what
the target is: a specific name (or a specific kind, if the target itself is a kind of thing) at the same
level as the target (if the target is a city, a city; if a building, a building; if an artwork, an artwork
title; if a person, a person). Generic descriptions ("a stadium", "an Indian city", "a man in a suit") are
not candidates. Names of other entities that are not offered as the target's identity are not candidates.
Give {"quote": "<the reasoning's exact words containing it, copied character for character including any *
or quotes, 5 to 15 words>", "candidate": "<the candidate>", "is_target": true or false}, or null if the
reasoning never proposes a candidate.
"target_mention": the FIRST place where the reasoning names the TARGET (or another name of it) as a possible
identity, even if it later rejects it. Give {"quote": "<exact words, 5 to 15 words>"}, or null if the
target is never named.
"turn_away": only if the target is named and the reasoning afterwards turns away from it (rejects it, doubts
it and settles on something else, or ends on something else): {"quote": "<exact words where it turns away,
5 to 20 words>"}; otherwise null.
"hints": for each hint, decide what the reasoning does with it. Judge the content of the hint only.
- "same": the reasoning states this hint (the same visual detail, in any wording, possibly less precisely);
- "different": the reasoning refers to the same object, person, text or region but reads, sees or
  identifies it differently;
- "omits": the reasoning mentions that object or region but never reports this detail, and says nothing
  that conflicts with it;
- "absent": the reasoning never refers to that object, person, text or region.
Give [{"id": "c1", "status": "same" or "different" or "omits" or "absent", "quote": "<the reasoning's first
words stating this detail (for same) or referring to it (otherwise), exact, at most 20 words, or empty>"},
...].
Output format: {"first_candidate": ..., "target_mention": ..., "turn_away": ..., "hints": [...]}

QUESTION: {q}
TARGET: {target}

HINTS:
{hints}

REASONING:
<<<
{think}
>>>
\end{verbatim}
\end{promptblock}
\caption{\textbf{Judge prompt for the clue mention rate.} The clue mention rate uses only the ``hints'' field of the output.}
\label{fig:supp_clue_judge}
\end{minipage}
\end{center}

\begin{center}
\begin{minipage}{0.98\textwidth}
\captionsetup{type=figure}
\begin{promptblock}[title={\textbf{Prompt}}]
\begin{verbatim}
A model was asked a question about the entity shown in an image. Below is its full response (reasoning,
then answer).

Entity shown in the image: {entity}{desc}{aliases}

1. Final: what does the model finally identify the image itself as showing? (This is about the image, not
   the answer to the question.)
   Gold = the entity above (any name or alias, or a part of it); Other = a different specific entity;
   None = never settles on a specific entity.
2. Mentioned: does the response name the entity above anywhere, even if it later rejects it? Yes or No.

Reply in exactly this form: Final: <Gold|Other|None>; Mentioned: <Yes|No>

Response:
{response}
\end{verbatim}
\end{promptblock}
\caption{\textbf{Judge prompt for Recognition Failure and Recognition Success.} \texttt{\{desc\}} is a short type description in parentheses, \eg, ``(place or building)'', and \texttt{\{aliases\}} is ``(also known as: \ldots)'' with the aliases of the entity.}
\label{fig:supp_recog_judge}
\end{minipage}
\end{center}

\begin{center}
\begin{minipage}{0.98\textwidth}
\captionsetup{type=figure}
\begin{promptblock}[title={\textbf{(a) Candidate generation} (GPT-6.1 Sol; image, question, and answer)}]
\begin{verbatim}
You will see an image, a question about it, and the correct answer.
List the visible cues that would let a person who knows the answer recall it, that is, name it from that
cue alone. A cue is either a characteristic detail of the answer itself, or something else in the image that
can be identified on its own (recognized, read, or matched to a known symbol) and points to the answer.
Do not list general descriptions that fit many entities of the same kind, for example "a smiling man with
short dark hair and brown eyes", "a tall church" or "a woman seated in a garden". Do not list facts that
cannot be seen, or anything not actually visible. Copy any writing exactly as written. Do not write the
answer's name inside a cue unless it is literally written in the image.
For each cue give "what" (at most 15 words) and "box" [x0, y0, x1, y1] as fractions of the image.
List as many as you can, most helpful for recall first. Return JSON only: {"cues": [ ... ]}

QUESTION: {q}
CORRECT ANSWER: {gold}
\end{verbatim}
\end{promptblock}
\begin{promptblock}[title={\textbf{(b) Hallucination check} (GPT-6 Luna; image and one zoomed crop per candidate, no answer)}]
\begin{verbatim}
You saw a full image and, for each observation below, a zoomed view of the region where it should be. For
each observation, decide whether it is visible in the image, looking at its zoomed view first and then the
full image:
"yes" if the main content is visible, even if the stated location or a minor detail of the description is
slightly off;
"no" if it is not there or is clearly different (for text, if the characters are different, answer "no" and
write what it actually says);
"unsure" if the image is too small or blurry to tell.
Return JSON only: {"checks": [{"id": "h1", "visible": "yes" or "no" or "unsure", "note": "<if not yes: what
is actually there, at most 12 words, else empty>"}, ...]}

OBSERVATIONS:
{obs}
\end{verbatim}
\end{promptblock}
\begin{promptblock}[title={\textbf{(c) Decisiveness check} (GPT-6 Luna; no image)}]
\begin{verbatim}
One detail seen in a photo is given below, together with the question asked about the photo.
From this detail and your knowledge, give the name of the specific entity the photo most likely shows.
Answer with the name only; if you cannot tell from this detail, answer "unknown".

QUESTION: {q}
DETAIL: {what}
\end{verbatim}
\end{promptblock}
\begin{promptblock}[title={\textbf{(d) Entity-answer judge}}]
\begin{verbatim}
A model was shown an image and asked: "{question}"

Entity actually shown in the image: {entity}{aliases}

Model's answer:
{answer}

Does the model's answer name the entity shown?
Gold = it names this entity (any of its names or aliases, or a clear part of the name such as the correct
surname or title).
Other = it names a different specific entity instead.
None = it does not commit to any specific entity (hedges, refuses, lists candidates without choosing, or
only describes what is visible).

Reply in exactly this form: Final: <Gold|Other|None>
\end{verbatim}
\end{promptblock}
\caption{\textbf{Prompts for constructing entity-recall clues.}}
\label{fig:supp_clue_prompts}
\end{minipage}
\end{center}

\clearpage
\twocolumn

\end{document}